\documentclass[10pt, a4paper]{article}

\usepackage{lrec-coling2024} % this is the new style
\usepackage{todonotes}
\usepackage{amssymb}
\usepackage{multirow}

\newcommand{\gspace}[0]{{\color{gray}\textvisiblespace}}

\title{mALBERT: Is a Compact Multilingual BERT Model Still Worth It?}

\name{Christophe Servan$^{1,2}$, Sahar Ghannay$^{1}$, Sophie Rosset$^1$} 

\address{$^1$Université Paris-Saclay, CNRS, LISN, $^2$QWANT\\
         \{firstname.lastname\}@lisn.upsaclay.fr\\}

\abstract{Within the current trend of Pretained Language Models (PLM), emerge more and more criticisms about the ethical and ecological impact of such models.
In this article, considering these critical remarks, we propose to focus on smaller models, such as compact models like ALBERT, which are more ecologically virtuous than these PLM. 
However, PLMs enable huge breakthroughs in Natural Language Processing tasks, such as Spoken and Natural Language Understanding, classification, Question--Answering tasks.
PLMs also have the advantage of being multilingual, and, as far as we know, a multilingual version of compact ALBERT models does not exist.
Considering these facts, we propose the free release of the first version of a multilingual compact ALBERT model, pre-trained using Wikipedia data, which complies with the ethical aspect of such a language model.
We also evaluate the model against classical multilingual PLMs in classical NLP tasks.
Finally, this paper proposes a rare study on the subword tokenization impact on language performances.
\\ \newline  }
\begin{document}

\maketitleabstract

\section{Introduction}

Recent advances in the field of Natural Language Processing (NLP) are due to the development of transfer learning and the availability of Pre-trained Language Models (PLM) based on Transformer architectures \cite{NIPS2017_3f5ee243}, such as BERT \cite{devlin-etal-2019-bert}.
As they provide contextualized semantic representation, they contribute both to advancing the state-of-the-art on several NLP tasks and also to evolving training practices through the use of fine-tuning.
%, they have not yet been widely adopted.

The recent trend consists of training large PLMs on ever larger corpora with an ever-increasing amount of parameters, which requires considerable computational resources that only a few companies and institutions can afford, such as GPT-4 \cite{GPT4OpenAI}, LLaMA \cite{touvron2023llama} or BLOOM \cite{scao2022bloom}. 
This trend raises questions about the temporal, financial, and environmental aspects of these models \cite{DBLP:conf/acl/StrubellGM19, sustainlp-2020-sustainlp}. 
Therefore, one of the favored tracks is the reduction of computational resources involved while pre-training, fine-tuning, and inference of these models.

As far as we know, compact models, such as the ALBERT model\cite{lan2019albert}, are a possible answer since they have been evaluated on the comprehension tasks covered by GLUE \citelanguageresource{wang2018glue} and the question-answering task with the SQuAD corpus \citelanguageresource{rajpurkar2016squad} with abundant data.
They also have shown their effectiveness on lower-scale learning problems in poorly endowed languages but only in a monolingual context \cite{lan2019albert, cattan-etal-2021-usability}. 
As far as we know, the multilingual version of such a model does not exist. 

All Pre-trained Language Model uses subword unit tokenization in order to alleviate the open vocabulary problem.
We take the opportunity of a new language model to conduct a short study of the impact of the subword unit vocabulary.
Subword units come from studies conducted in machine translation using compression methods in order to reduce the vocabulary amount and to handle the Out-Vocabulary-Words \cite{ chitnis2015variable,schuster2012japanese,wu2016google,sennrich2016neural,kudo2018sentencepiece}.

These subword unit approaches are linguistic-free and are mainly models, which are estimated on raw text.
On the other side, it has been observed that these subword models do not correspond to linguistic units such as morphemes, affixes, etc. \cite{huck2017target,machavcek2018morphological,ataman2017linguistically,pinnis2017neural}.
%However, these approaches were not able to beat linguistic-free approaches.

In order to conduct our subword comparison, we create three versions of the same ALBERT model, which are trained on the same data but with different tokenization. 
%The goal of this subword study is to detect if there are significant impacts
The goal of this subword study is to verify the impacts of subword models associated to our ALBERT models in NLP tasks.
Especially in tokens class classification tasks, such as Named Entity Recognition or Spoken Language Understanding tasks. 
% Several previous studies were conducted, especially in the field of machine translation, whci lead.

\textbf{Contributions:} First, this paper presents the release of a multilingual version of ALBERT \cite{lan2019albert}: mALBERT\footnote{ \url{https://huggingface.co/cservan/malbert-base-cased-32k}

\url{https://huggingface.co/cservan/malbert-base-cased-64k}

\url{https://huggingface.co/cservan/malbert-base-cased-128k}}
% \footnote{ \url{https://huggingface.co/XXXXXXXX}}
, trained on open-source and ethical data;
second, we propose a study of the subword tokenization process focused on the vocabulary size impact. 
We also measure the tokenization impact, which is correlated with the subwords segmentation rate of tokens.

The paper is organized as follows: first, we present the model architecture and the pre-training details in Section \ref{sec:pretrain}; Section \ref{sec:XP} details the experiments conducted using our new models and the tokenization study; Finally, the last section presents the conclusion and outcomes of this paper.

%Experiments show the great efficiency of our model, and the tokenization impact is bla bla bla 
% \remcs{A compléter pour la fin}

% blabla
\section{Model Pre-training}
\label{sec:pretrain}

As far as we know, there is no multilingual compact model. 
We therefore propose to pre-train a new version of ALBERT from scratch: \textit{mALBERT}.

ALBERT is based on parameter sharing/reduction techniques that enable us to reduce the computational complexity and speed up training and inference phases.
Compared to previous compact models such as DistilBERT \cite{DBLP:journals/corr/abs-1910-01108}, Q-BERT \cite{DBLP:conf/aaai/ShenDYMYGMK20} or TernaryBERT \cite{zhang2020ternarybert}, ALBERT is to the date the smallest pre-trained models with 12 million parameters and \textless50 megabyte model size.
ALBERT models also show their ecological advantages regarding bigger models \cite{cattan2022benchmarking}.

\subsection{Data}

\begin{figure*}
    \centering
    \includegraphics[width=0.75\linewidth]{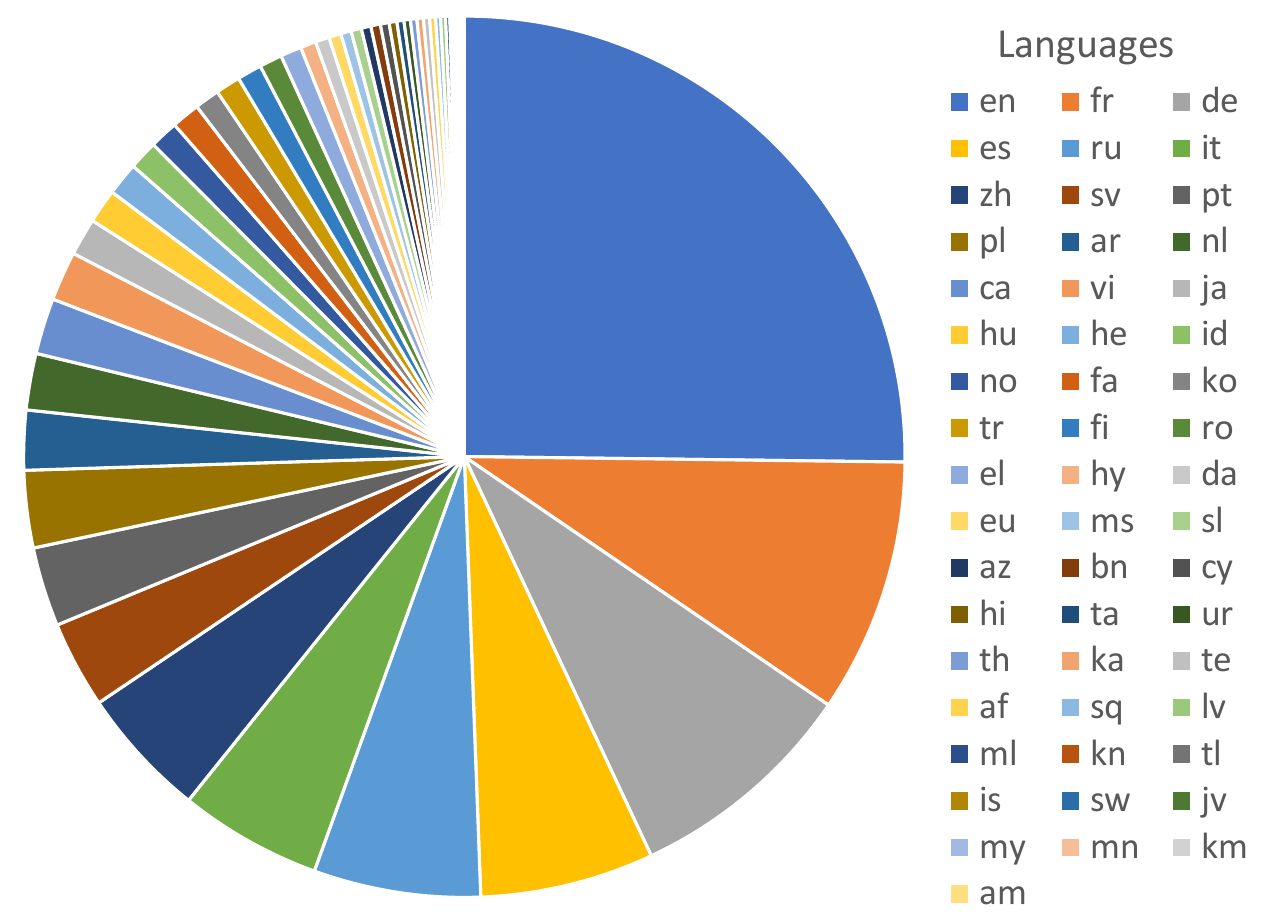}
    \caption{Language distribution (52 languages) over the training corpus. In the legend, languages are presented according to their representativity: from left to right and from up and down. The most representative language is English (\textit{en}) and the least one is Amharic (\textit{am}) }
    \label{fig:enter-label}
\end{figure*}

Aiming to use open-source and ethical data to pre-train the mALBERT model, we decided to use only Wikipedia data for each language.
%In order to do so, we collect Wikipedia corpora for each considered language. 
Figure \ref{fig:enter-label} presents the language distribution of the Wikipedia corpus collected on January 2023.
The corpus is roughly 21 billion words across 50 most common languages on Wikipedia, plus English and Basque.
%This represents roughly 21 billion words. 

As for many other multilingual models, English prevails the whole corpus with French, German, and Spanish. 
These four languages represent nearly 50\% of the corpus.

\subsection{Subword unit}

%All Pre-trained Language Model uses subword unit tokenization in order to alleviate the open vocabulary problem.
The subword unit tokenization model chosen for our multilingual ALBERT model is based on a unigram language model approach \cite{kudo2018sentencepiece}. 
This subword unit approach was chosen because it enables us to fix the final amount of vocabulary.

Three subword unit models were trained on a subpart of the corpus selected randomly, in order to study the impact of the tokenization process on the final ALBERT model performances.
%This unsupervised tokenizer implements subword units by using a unigram language model, which is an effective way to alleviate the open vocabulary problems in neural networks.
The tokenization models differ only with the amounts of the final vocabulary generated: 32k, 64k, and 128k.

% \remcs{Description des données et des traitements appliqués}
% \begin{itemize}
%     \item OPUS \cite{tiedemann2012opus}
%     \item Wikipedia 
%     \item Sentencepiece \cite{kudo2018sentencepiece}
% \end{itemize}

\subsection{Training parameters}

%French super calculator JeanZay

Models are trained for roughly 9000 hours on the ANONYMIZED CALCULATOR NAME, using the UER-py toolkit \cite{zhao2019uer} jointly with DeepSpeed \cite{rasley2020deepspeed}, and use multiple training objectives (masked language modeling and next sentence or sentence order prediction).
We use the same learning configuration as the original model with a batch size of $128$ and an initial learning rate set to $3.125\times10\textsuperscript{-4}$.

Finally, we pre-train three models based on the same amount of corpus, with the same amount of parameters, but they differ only by the amount of input vocabulary.
We noted our final models as follows: \textit{mALBERT-128k}, \textit{mALBERT-64k}, and \textit{mALBERT-32k}, which respectively use an amount of 128k, 64k, and 32k tokens.
% Our experiments are also based on the large multilingual model mBERT \cite{devlin-etal-2019-bert} as well as on the compact multilingual models with a distilled version of mBERT: Distil-mBERT \cite{DBLP:journals/corr/abs-1910-01108}.

% \remcs{Détails des hyper-paramètres}

% \begin{itemize}
%     \item Toolkit UER-py (torch) \cite{zhao2019uer}
%     \item Batch size etc.
% \end{itemize}

    \begin{table*}[h!]
        \centering
\resizebox{\textwidth}{!}{
        \begin{tabular}{lccccccccccc}
\hline
Models $\diagdown$ Tasks & MMNLU &   MultiATIS++ &  CoNLL2003  & MultiCoNER & SNIPS &  MEDIA \\
\hline
% Models & F1-mesure & intervalle  &  &  &  & F1-mesure & intervalle  &  &  &  &  &  & \\
% \hline
mBERT & 73.46$^*$ (0.11) &  92.22 (0.11)  & 95.59$^*$ (0.10) & 66.36$^*$ (0.18)  & 96.09 (0.31) & 87.90$^*$ (0.09) \\
Distil-mBERT & 72.44$^*$ (0.08) & 91.69 (0.09) & 94.59$^*$ (0.13)  & 61.26$^*$ (0.13) & 94.95 (0.22)& 86.83$^*$ (0.21) \\
EnALBERT & N/A & N/A  & 89.67$^*$ (0.34) & 42.36$^*$ (0.22)  & 95.95   (0.13) & N/A  \\
FrALBERT & N/A & N/A  & N/A  & N/A    & N/A & 81.76 (0.59) \\
mALBERT-128k & 65.81$^*$ (0.11) &  89.14 (0.15) & 88.27$^*$ (0.24)  & 46.01$^*$ (0.18) & 91.60 (0.31) & 83.15$^*$  (0.38) \\
mALBERT-64k & 65.29$^*$ (0.14) &  88.88 (0.14)  & 86.44$^*$ (0.37) & 44.70$^*$ (0.27) & 90.84 (0.47) &  82.30 (0.19)\\
mALBERT-32k & 64.83$^*$ (0.22)& 88.60 (0.27)  & 84.96$^*$ (0.41)  & 44.13$^*$ (0.39)  & 89.89 (0.68)& 82.04 (0.28)\\
\hline
        \end{tabular}
        }
        \caption{Results on several slot-filling tasks regarding the F1-measure score. The results are the mean of 10 different runs, and the standard deviation is noted between parenthesis. $^*$: p-value $< 0.05$.}
        \label{tab:POS}
        \vspace*{-0.4cm}
    \end{table*}

\begin{table*}[]
    \centering
    \begin{tabular}{lcccc}
    \hline
Models $\diagdown$ Tasks & MMNLU & MultiATIS++ & SNIPS & SST2 \\
    \hline
mBERT & 80.32$^*$ (0.09) & 96.14$^*$ (0.17) & 97.31   (0.31) & 46.49$^*$   (0.76) \\
Distil-mBERT & 78.23$^*$ (0.08) & 92.79$^*$ (0.35) & 97.69   (0.25) & 43.59   (0.31) \\
EnALBERT &  N/A & N/A  &  97.60 (0.11) & 43.66   (1.88)\\
mALBERT-128k & 72.35$^*$ (0.09) & 90.58 (0.98) & 96.84   (0.49)  & 34.66   (1.46) \\
mALBERT-64k & 71.26$^*$ (0.11) & 90.97 (0.70) & 96.53   (0.44) & 34.64   (1.02) \\
mALBERT-32k & 70.76$^*$ (0.11) & 90.55 (0.98) & 96.49   (0.45)  & 34.18   (1.64) \\
    \hline
    \end{tabular}
    \caption{Results on several classification tasks regarding the Accuracy score. The results are the mean of 10 different runs, and the standard deviation is noted between parenthesis. $^*$: p-value $< 0.05$.}
        \label{tab:classif}
\end{table*}

\section{Experiments}
\label{sec:XP}
% \remcs{2 tâches: tagging labelization (+NER) voir si je peux ajouter squad...}
Our three new models are benchmarked on two kinds of classical NLP tasks: the slot-filling and classification tasks.
These tasks use standard fine-tuning approaches, in which fine-tuning and evaluation scripts are provided by HuggingFace \cite{wolf2019huggingface}. 
%The joint fine-tuning approach uses the jointBERT approach proposed by \citet{chen2019bert}.
For each experiment, we do not seek to have the best score, but a point of comparison for our models.

We compare our new multilingual ALBERT model to the large multilingual model \textit{mBERT} \cite{devlin-etal-2019-bert} as well as on the compact multilingual models with a distilled version of mBERT: \textit{distil-mBERT} \cite{DBLP:journals/corr/abs-1910-01108}.
Our comparison includes also some monolingual versions of ALBERT for English (noted \textit{EnALBERT} in CoNLL2003 and MultiCoNER tasks) and French (in MEDIA), noted \textit{FrALBERT} \cite{cattan-etal-2021-usability}.
% Our experiments are also based on the large multilingual model mBERT \cite{devlin-etal-2019-bert} as well as on the compact multilingual models with a distilled version of mBERT: Distil-mBERT \cite{DBLP:journals/corr/abs-1910-01108}.
%The mBERT and Distil-mBERT are fine-tuned through 10 epochs, and compact models are fine-tuned through 40 epochs.
%This difference came from we observe a need of more training time in order to have 

Finally, we do not compare our models with bigger LLM such as GPT-4 \cite{GPT4OpenAI}, LLaMA \cite{touvron2023llama} or BLOOM \cite{scao2022bloom}, for resources and ecological considerations. 

\subsection{Slot-filling benchmark}

Six slot-filling tasks are used to benchmark our new mALBERT models: two multilingual understanding tasks, Massively Multilingual NLU 2022 (MMNLU) \citelanguageresource{fitzgerald2022massive} and MultiATIS++ \citelanguageresource{xu-etal-2020-end};  two Named Entity Recognition monolingual tasks: CoNLL2003 \citelanguageresource{tjong-kim-sang-de-meulder-2003-introduction} and MultiCoNER \citelanguageresource{malmasi-etal-2022-semeval}; and two monolingual language understanding tasks: SNIPS \citelanguageresource{coucke2018snips} and MEDIA \citelanguageresource{bonneau2009media}. 

%We compare our new multilingual ALBERT model to the large multilingual model mBERT \cite{devlin-etal-2019-bert} as well as on the compact multilingual models with a distilled version of mBERT: distil-mBERT \cite{DBLP:journals/corr/abs-1910-01108}.
%We also compare our models to some monolingual versions of ALBERT available for English (in CoNLL2003 and MultiCoNER tasks) and French (in MEDIA), noted FrALBERT \cite{cattan-etal-2021-usability}.

Table \ref{tab:POS} presents the results obtained on the slot-filling tasks according to the F1-measure. 
For every task and model, we perform 10 runs with a different seed each time.
Over all tasks, \textit{mBERT} and \textit{Distil-mBERT} obtain the best results.
On the one hand, monolingual ALBERT models perform better on CoNLL2003 and SNIPS tasks. 
On the other hand, one can observe that \textit{mALBERT} models perform better than \textit{FrALBERT} and \textit{EnALBERT} models on MultiCoNER and MEDIA tasks, respectively.
This ensure us that \textit{mALBERT} is comparable with other monolingual ALBERT models.

In all slot-filling tasks, the 128k version of the mALBERT model performed better than the two other variants. 
Moreover, we observe a hierarchy in the mALBERT model versions according to their vocabulary size: the one with the smaller vocabulary is the worst and the 64k mALBERT variant is second.

\subsection{Classification benchmark}
For the classification benchmark, we evaluate our models against four tasks.
First, two multilingual tasks: Massively Multilingual NLU 2022 (MMNLU)  \citelanguageresource{fitzgerald2022massive} and MultiATIS++ \citelanguageresource{xu-etal-2020-end} ; second, two monolingual tasks: SNIPS \citelanguageresource{coucke2018snips} and Stanford Sentiment Treebank v2 (SST2) \citelanguageresource{socher2013recursive}.

Like in the slot-filling task, bigger models obtained the best results over all tasks.
Focusing on the mALBERT models, they obtained results with a $p-value$ lower than 0.05 only in the MMNLU task.
On other tasks (MultiATIS++, SNIPS, and SST2), the significativity of results between our new models are not reached.
Considering MMNLU results, this leads us to the same observation we have with the slot-filling task: the mALBERT model performances are ranked according to their vocabulary size.

% \begin{itemize}
%     \item STSA+B (classif tweets) \remcs{Moi ?}
% \end{itemize}
% \subsection{Joint slot-filling and classification task}
% \remcs{ça dépendra des résultats}
% % \begin{itemize}
% %     \item SNIPS  \remcs{Sahar ?}
% %     \item ATIS  \remcs{Sahar ?}
% %     \item Massive  \remcs{Sahar}
% % \end{itemize}

%     \begin{table*}[]
%         \centering
%         \begin{tabular}{lcccccccc}
% \hline
% Models $\diagdown$ Tasks & Massive\_all &   MultiATIS & snips \\
% \hline
% % Models & F1-mesure & intervalle  &  &  &  & F1-mesure & intervalle  &  &  &  &  &  & \\
% % \hline
% mBERT &79.4/64.9  & 98.2/95.4&  98.8/96.8\\
% Distil-mBERT & 81.5/72.1 & 98.1/95.3 &98.4/95.5\\

% mALBERT-128k &79.2/64.3 & 93.9/87.9  &97.7/90.7\\

% mALBERT-64k &79.7/65.3 &  93.6/87.8 &97.7/90.4 \\

% mALBERT-32k &79.5/65.0 & 93.6/87.6  &89.0/90.3\\

% \hline
%         \end{tabular}
%         \caption{Results on joint intent and several slot-filling tasks regarding $intent\_acc$/F-measure.}
%         \label{tab:my_label}
%     \end{table*}

\begin{table*}[]
    \centering
\resizebox{\textwidth}{!}{
    \begin{tabular}{lcccccccccccc}
    \hline
\textbf{Plain text} & acquisition & of & Daniels & Pharmaceuticals & Inc & of & St. & Petersburg & , & Fla. \\
%    \hline
\textbf{Reference} & O & O & B-ORG & I-ORG & I-ORG & O & B-LOC & I-LOC & O & B-LOC \\
   \hline
\textbf{Tok-32k}  & \textbf{\_}acquis\gspace{}i\gspace{}tion  & \textbf{\_}of &  \textbf{\_}Daniel\gspace{}s  & \textbf{\_}Pharmac\gspace{}e\gspace{}u\gspace{}tical\gspace{}s  & \textbf{\_}Inc  & \textbf{\_}of  & \textbf{\_}St\gspace{}.  & \textbf{\_}Petersburg  & \textbf{\_},  & \textbf{\_}F\gspace{}la\gspace{}.   \\
\textbf{mALBERT-32k} & O & O & B-ORG & I-ORG & B-ORG & I-ORG & I-ORG & I-ORG & O & B-ORG  \\
    \hline
\textbf{Tok-64k}  & \textbf{\_}acquisition  & \textbf{\_}of  & \textbf{\_}Daniel\gspace{}s  & \textbf{\_}Pharmac\gspace{}e\gspace{}u\gspace{}tical\gspace{}s  & \textbf{\_}Inc  & \textbf{\_}of  & \textbf{\_}St\gspace{}.  & \textbf{\_}Petersburg  & \textbf{\_},  & \textbf{\_}Fla\gspace{}. \\
\textbf{mALBERT-64k} & O & O & B-PER & I-PER & B-ORG & O & B-ORG & I-ORG & O & B-PER \\
    \hline
\textbf{Tok-128k} & \textbf{\_}acquisition  & \textbf{\_}of  & \textbf{\_}Daniel\gspace{}s  & \textbf{\_}Pharmaceutical\gspace{}s  & \textbf{\_}Inc  & \textbf{\_}of  & \textbf{\_}St\gspace{}.  & \textbf{\_}Petersburg  & \textbf{\_},  & \textbf{\_}Fla\gspace{}.  \\
\textbf{mALBERT-128k} & O & O & B-ORG & I-ORG & I-ORG & O & B-LOC & I-LOC & O & B-LOC \\
    \hline

    \end{tabular}
}
\caption{Example of segmentation / tokenization for each model and the label detected by the model for the CoNLL2003 task (NER). In this table the original input text is noted \textit{Plain text}, with its gold labelization (\textit{Reference}). Then each next row corresponds to a tokenization model (\textit{Tok-32k}, \textit{Tok-64k}, \textit{Tok-128k}) and the output of the associated model (\textit{mALBERT-32k}, \textit{mALBERT-64k}, \textit{mALBERT-128k}). The token segmentation in subwords is indicated with a special character as separator (\gspace{}). }
    \label{tab:exsegtag}
            \vspace*{-0.4cm}
\end{table*}

\begin{table}[]
    \centering
\resizebox{\linewidth}{!}{
    \begin{tabular}{lrrr}
\hline
    \textbf{Subword}  & \multirow{2}{*}{Tok-32k~~} & \multirow{2}{*}{Tok-64k~~} & \multirow{2}{*}{Tok-128k~~} \\
    \textbf{vocab. size } & & &\\
\hline
         \textbf{NE} & 120.59 \% & 85.28 \% &  62.69 \% \\
         \textbf{Not NE}& 57.64 \% & 36.04 \% & 25.37 \% \\
\hline
    \end{tabular}
    }
    \caption{Impact of the tokenization on word type (i.e.: belong to a Named Entity or not.) in the CoNLL2003 task. We reported the percentage of additional segmentation observed.}
    \label{tab:vocabimpact}
            \vspace*{-0.4cm}
\end{table}

\subsection{Tokenization Impact}

The starting point of this study is to measure the impact of the tokenization of subword unit models.
We compare our three tokenization models: 32k, 64k, and 128k codes (which, in our case, corresponds to the final amount of vocabulary).
This study focuses on a Named Entity task, the CoNLL2003 task, which is a slot-filling task based on a token classification method.
This means the segmentation of the token in subwords could increase the sentence context, which may impact the final labelization result.

%In order to \remcs{... je trouve pas...}
In order to measure the possible impact of the subword tokenization, we estimate the amount of additional segmentation according to Name Entity (NE) labels (table \ref{tab:vocabimpact}). 
We can observe a significant impact on the token  segmentation associated with NE:
the 128k subword model segmentation of tokens produces 62\% more subwords, meanwhile, the 32k subword model produces 120\% of additional subwords.

We push deeper into the analysis and estimate the Pearson correlation score between the segmentation of the word in subwords and the non-detection of the associated label of the original token. 
The correlation score is 0.44, which implies a moderate correlation of the tokenization impact on the labelization process.
This means the more the entity is segmented, the less accurate the model is to identify the right entity.

Table \ref{tab:exsegtag} presents an example of the segmentation and labelization of the sequence ``acquisition of  Daniels Pharmaceuticals Inc of St. Petersburg, Fla.''. 
In this example, we can observe that the most split word is ``Pharmaceuticals''. 
This sequence of subwords illustrates the impact, especially on the label of next word `Inc''.
The impact can directly be observed on the label of words ``Pharmaceuticals'' and ``Fla.''. 
The right labelization is obtained once the whole segment is the less splitted in subwords.

The subword tokenization seems to interfere with the labelization of these tokens made by the model.
Finally, these remarks on subword tokenization seem obvious. 
Still, as far as we know, we have not found any study on the impact of tokenization on pre-trained language models.
This first study shall be pushed further to precisely measure the impact of subword tokenization models on other tasks and domains.
%in two ways: locally and remotely concerning respectively the current token and the surrounding tokens.
% It seems well labeled in the 32k model, but it seems to impact the next words in this example.
% For instance, the word ``Inc'' is correctly labeled only using the 128k model.

% \begin{table*}[]
%     \centering
% \resizebox{\textwidth}{!}{
%     \begin{tabular}{lcccccccccccc}
% \hline
% Plain text & acquisition & of & Daniels & Pharmaceuticals & Inc & of & St. & Petersburg & , & Fla. & , \\
% \hline
% 32k  & \textbf{\_}acquis i tion  & \textbf{\_}of &  \textbf{\_}Daniel s  & \textbf{\_}Pharmac e u tical s  & \textbf{\_}Inc  & \textbf{\_}of  & \textbf{\_}St .  & \textbf{\_}Petersburg  & \textbf{\_},  & \textbf{\_}F la .  & \textbf{\_}, \\
% 64k  & \textbf{\_}acquisition  & \textbf{\_}of  & \textbf{\_}Daniel s  & \textbf{\_}Pharmac e u tical s  & \textbf{\_}Inc  & \textbf{\_}of  & \textbf{\_}St .  & \textbf{\_}Petersburg  & \textbf{\_},  & \textbf{\_}Fla .  & \textbf{\_}, \\
% 218k & \textbf{\_}acquisition  & \textbf{\_}of  & \textbf{\_}Daniel s  & \textbf{\_}Pharmaceutical s  & \textbf{\_}Inc  & \textbf{\_}of  & \textbf{\_}St .  & \textbf{\_}Petersburg  & \textbf{\_},  & \textbf{\_}Fla .  & \textbf{\_}, \\
% \hline
%     \end{tabular}
% }
%     \caption{Example of segmentation / tokenization for each model.}
%     \label{tab:exseg}
% \end{table*}

% \remcs{Proposition de décortiquer les résultats et la segmentation des sous-unités lexicales}

% \subsection{Ecological impact}
% \remcs{Classique avec CodeCarbon}

\section{Conclusion}

This paper presents the first multilingual ALBERT model (mALBERT), pre-trained on Wikipedia dump in 52 languages.
The model comes with three vocabulary size variants: 32k, 64k, and 128k.
All variants were pre-trained on data extracted from 91 Go of Wikipedia dumps, which represents more than 21 billion words. 

So, is a multilingual compact still worth it? Evaluations in classical NLP tasks (slot-filling and classification tasks) show the multilingual version of ALBERT has comparable results to the monolingual versions used in this paper. From an ecological and resource aspect, one model pre-training on GPU time took 9k hours, which is far from the million hours for the BLOOM LLM.

%So, is a multilingual compact still worth it? According to the quality aspect of the mALBERT compared to other ALBERT models, yes. For all ecological aspects already shown by \citet{cattan2022benchmarking}, yes again.

The tokenization study, focused on vocabulary size, gives some feedback about the importance of the impact of subword tokenization. 
The moderate correlation observed on a classical Named Entity task enables us to say the more you split tokens into subwords, the less the Entity is well detected.

In the next steps, the extension of the subword tokenization model study will investigate which kind of segmentation could be the best for Pre-trained Language Models on more NLP tasks. 

The three versions of the model are freely available on huggingFace\footnote{ \url{https://huggingface.co/cservan/malbert-base-cased-32k}

\url{https://huggingface.co/cservan/malbert-base-cased-64k}

\url{https://huggingface.co/cservan/malbert-base-cased-128k}} 
\section{Acknowledgements}

This paper was funded by the \textit{Multiligual SLU for Contextual Question Answering (MuSCQA)} project from "France Relance" of French government funded by French National Research Agency (ANR), grant number: ANR-21-PRRD-0001-01.
This work was performed using HPC resources from GENCI–IDRIS (Grant 2023-A0131013834).

% \nocite{*}
\section{Bibliographical References}\label{sec:reference}

\bibliographystyle{lrec-coling2024-natbib}
\bibliography{lrec-coling2024}

\section{Language Resource References}
\label{lr:ref}

\bibliographystylelanguageresource{lrec-coling2024-natbib}
\bibliographylanguageresource{languageresource}

\end{document}